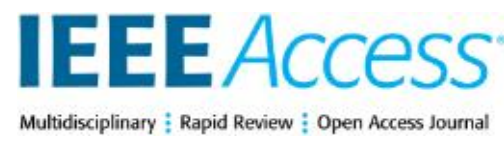

# Evaluation of Conversational Agents: Understanding Culture, Context and Environment in Emotion Detection

**Martha T. Teye[1,2], Yaw Marfo Missah[2], Emmanuel Ahene[2] Twum Frimpong[2] and Auxane Boch[3]**

[1]Cluster of Excellence, University of Stuttgart, 70174 Stuttgart, German
[2]Department of Computer Science, Kwame Nkrumah University of Science and Technology, Kumasi, Ghana
[3]Institute for Ethics in Artificial Intelligence, Technical University of Munich, 80333 Münich, Germany

Corresponding author: Martha Teiko Teye (e-mail: martha.teye9@gmail.com).

**ABSTRACT** Valuable decisions and highly prioritized analysis now depend on applications such as facial biometrics, social media photo tagging, and human robots interactions. However, the ability to successfully deploy such applications is based on their efficiencies on tested use cases taking into consideration possible edge cases. Over the years, lots of generalized solutions have been implemented to mimic human emotions including sarcasm. However, factors such as geographical location or cultural difference have not been explored fully amidst its relevance in resolving ethical issues and improving conversational AI (Artificial Intelligence). In this paper, we seek to address the potential challenges in the usage of conversational AI within Black African society. We develop an emotion prediction model with accuracies ranging between 85% and 96%. Our model combines both speech and image data to detect the seven basic emotions with a focus on also identifying sarcasm. It uses 3-layers of the Convolutional Neural Network in addition to a new Audio-Frame Mean Expression (AFME) algorithm and focuses on model pre-processing and post-processing stages. In the end, our proposed solution contributes to maintaining the credibility of an emotion recognition system in conversational AIs.

## I. INTRODUCTION

The area of Human-Artificial Intelligence (AI) interactions research considers the psychological and social interactions, as well as the methods used to facilitate engagement between AIs and humans as though it were a human-to-human interaction. Over the years, conversational AI has evolved and become more useful to daily work. Since part of the conversational AI role and competence is to seamlessly emulate the likeness of humans, the relationship between the two should present lesser gaps and ethical concerns [1]. These interactions are usually achieved by the use of social robots[2] which are AI systems designed to interact with humans and other robots whiles conforming to rules attached to social behaviours.

In more recent times, one of the most sought-after AI technologies is chatbots, which have been integrated on many software platforms. From Facebook's platform which allows developers to create bots, to Slack's Kip, which helps to order office supplies right from your conversational channel with your co-workers [3]. Today, conversational AIs are used by a large part of the population: the world population uses 3.25 billion digital voice assistants in 2019 [4]. Notably, in this COVID-19 season where it has become necessary for less human-to-human interactions, AI's come in handy for educational institutions, entertainment, and trade.

Our current society cannot do away with the exponential growth in the fields of intelligence and technology [5]. Unfortunately, black people fall in the minority when it comes to technologies such as facial recognition. However, the risk of not taking opportunities and not using technologies due to fear seems to grow, especially in most parts of Africa [6]. Since the technologies would keep evolving it is best to intensify knowledge into their creations, use them and act to mitigate their risks [7].

It is necessary to make sure AI applications are guided to fit into the cultural and ethical values of the future society [8]. Questions arise as to how AI-based software be regulated and certified based on compatibility with social and technical norms. Our paper focuses on the concerns that are being raised about the interactions between humans and AIs and the current and future needs of emotion recognition. This involves understanding the role that it plays in our future society considering all stakeholders and societal factors to be able to cater for regulatory issues that may arise.

We highlight the major gaps in Human-Computer Interactions with regards to emotion detection in Section 2. We then focus on mitigating some of the biases that occur during the development and deployment of the machine learning models. We provide a better model which focuses on model pre-processing and post-processing stages by incorporating an Audio-Frame Mean Expression. The key methods used in building the Convolutional Neural Network ML model for both image and audio data are described in Section 3. This includes data balancing, pre-processing, and augmentation processes used to mitigate bias in the model. Section 4 provides detailed discussions and explanations based on the results from section 3. It also justifies the reasons for considering image and audio datasets and highlights the importance of the new approach used in mitigating bias. Finally, a summary of the entire paper, findings on emotion detection, and recommendations for future work are captured in section 5.

## II. RELATED WORK

Emotions are one of the key drivers in human social life [9]. Humans tend to develop emotions for human and sometimes non-human entities, including robots and interactional systems. By this, scientists began designing and creating robots to assume the physical embodiment and image of humans' nature. Constant interaction between AI and humans seems to create a special attachment between the two parties. Expressions of some form of emotions could further lead to emotional trust in human-AI interactions. Ekman [10] proposed six major emotions of interest in modelling AI's, which include fear, anger, disgust, happiness, sadness, and surprise. In 2017, a prototype was created to detect these major emotions including sarcasm from posts made by Facebook users [11]. The hybrid approach which included using the substantivized terms, similarity measure, polarity correction, and pattern analysis allowed for the detection of emotions with grammatically correct sentences.

Large image and text-based datasets made available from companies such as Microsoft, IBM, Reddit, and Twitter have been used in several Machine Learning research to create diversity and help improve existing models. An instance is the automatic tagging of prosodic phrases which are segments of speech that occur with a single pitch was employed to extract natural language data from real-life conversations using the support vector machine model [12]. This approach aided in the segmentation of words and phrases including non-verbal content. Achieving a 52% accuracy using this Natural Language Processing (NLP) model, the result showed that even with the seven major emotions, non-verbal sounds played a crucial part in achieving high accuracies for emotion detections.

Faces or facial expressions have also been one major and effective way of detecting emotions. It exposes the state of mind of an individual and could also help in determining other social and mental characteristics or responses [13]. When considering only speech emotion recognition, physiological stress response seems to be key in order not to misunderstand the intentions of the speech. Acculturative stress can also be seen as similar among people of the same cultural group [14]. Moreover, since facial expressions are also often similar among people of the same cultural group, it is usually easier for understanding and interpretation of emotions shown by individuals of similar cultures.

A new system model took into consideration two people of a small social group to identify the patterns that affect nonverbal interactions [15]. In doing this, the system did not only take into consideration the characteristics of the individual but also paid particular attention to the social environment where this interaction is being held [15]. The approach goes beyond reliance on only the thoughts, emotions, and influence of a person's interaction. Hence, supporting that ecological factors are provided by some psychological effects which affect emotion recognition.

However, with regards to macro and micro expressions, most standalone facial expression recognition systems tend to perform well in identifying macro expressions. These are expressions that can be noticed on a large area of a person's face and are visible for about 0.5 and 4 seconds [16]. Unlike micro-expressions, which happen spontaneously (usually between 65-500 microseconds), macro expressions exhibit facial structural movement which can be easily noticed. This usually causes attentional bias and may be overlooked in the process of designing conversational AIs [17].

Puri et al. [18] implemented the Open-Source Computer Vision Library (OpenCV) to detect faces and predict the same 7 major facial expressions. The OpenCV library has over 2000 machine learning image processing algorithms. However, the detection algorithm only worked best on images that have been cropped to the size and properties of training data and this could not be used to process real-time data. Similarly, a comparison using autoencoders [19] and an 8-layered CNN algorithm showed that by performing relevant hyperparameter tuning, CNN models [20]–[22] provided better results.

AI Fairness 360 was also introduced by IBM to help mitigate some algorithmic and data biases in building AI models [23]. It tests for the possible causes of biases in data or AI models and provides recommendations for mitigating such biases. Most of the bias mitigation could be achieved during various processes when building the machine learning pipelines. This includes the during, the pre-processing, in-processing, and post-processing stages. The dataset is biased on the underrepresented group (black demographics) in 4 out of 5 metrics computed. Using re-weighing mitigation techniques to improve the dataset, reduced the bias to acceptable levels.

Recently, several CNN architectures [20]–[22] were experimented on including Google's Inception-v3, ResNet 18, 34 and 50, VGG-16, and DenseNet-161. The models also considered datasets that showed several profiles of the face aside from the popular frontal image. The results showed that DenseNet-161 performed better with an accuracy between 96.51 and 99.52 per cent [24].

Now the models described above tend to focus only on either static images or speech which works well for basic emotion detections with little focus on diversity, micro-expressions, and real-time detection [2]. Our solution focuses on maintaining the credibility of an emotion recognition system in conversational AIs. The Convolutional Neural Network (CNN) has been implemented on both facial expression recognition (FER) [25] and speech emotion recognition (SER) datasets [26]–[29] to be able to cater for tests on video frames. In addition to the CNN implementation, we introduce a new mathematical approach, termed the Audio-Frame Mean Expression (AFME) to validate and improve the result from the neural network.

## III. MODEL DESIGN

We combine the Facial Expression dataset and randomly collected data from social media pages of African descent and perform rigorous data pre-processing to balance the data representation and identify relevant features for training the machine learning model. This model is then trained using the Long Short-Term Memory (LSTM) and Convolutional Neural Networks (CNN) with different hyperparameter values. Further, we propose a new mathematical model which is incorporated to provide better accuracies for all emotions described as shown in Fig. 1.

### A. ENVIRONMENT SETUP

Our model is built using the Python Programming Language (Python 3) in the Jupyter notebook (V2.1.5) environment with libraries such as Numpy, Pandas, and Matplotlib.

### B. DATASET

Our model depends on two main sources of data to build and train the machine learning models. The Facial Emotion Recognition datasets (FER, 2013) and the Speech Emotion Recognition datasets (SER). The sources of SER datasets were:

- Crowd-sourced Emotional Multi-modal Actors Dataset - CREMA-D [26].
- Ryerson Audio-Visual Database of Emotional Speech and Song - RAVDESS [27].
- Surrey Audio-Visual Expressed Emotion – SAVEE [28].
- Toronto emotional speech set - TESS [29]

Apart from images gathered from the FER dataset, we extracted 1000 extra locally generated Ghanaian images. This data includes 75 happy, 145 surprises, 150 fear, 150 angry, 75 sad, 75 neutral, and 320 disgust faces with the final representation as shown in Fig. 2. Sample images that were gathered to complement the open-source data are shown in Fig. 4. This approach is to make sure the algorithm understands some expressions which are peculiar to the African (Ghanaian Community).

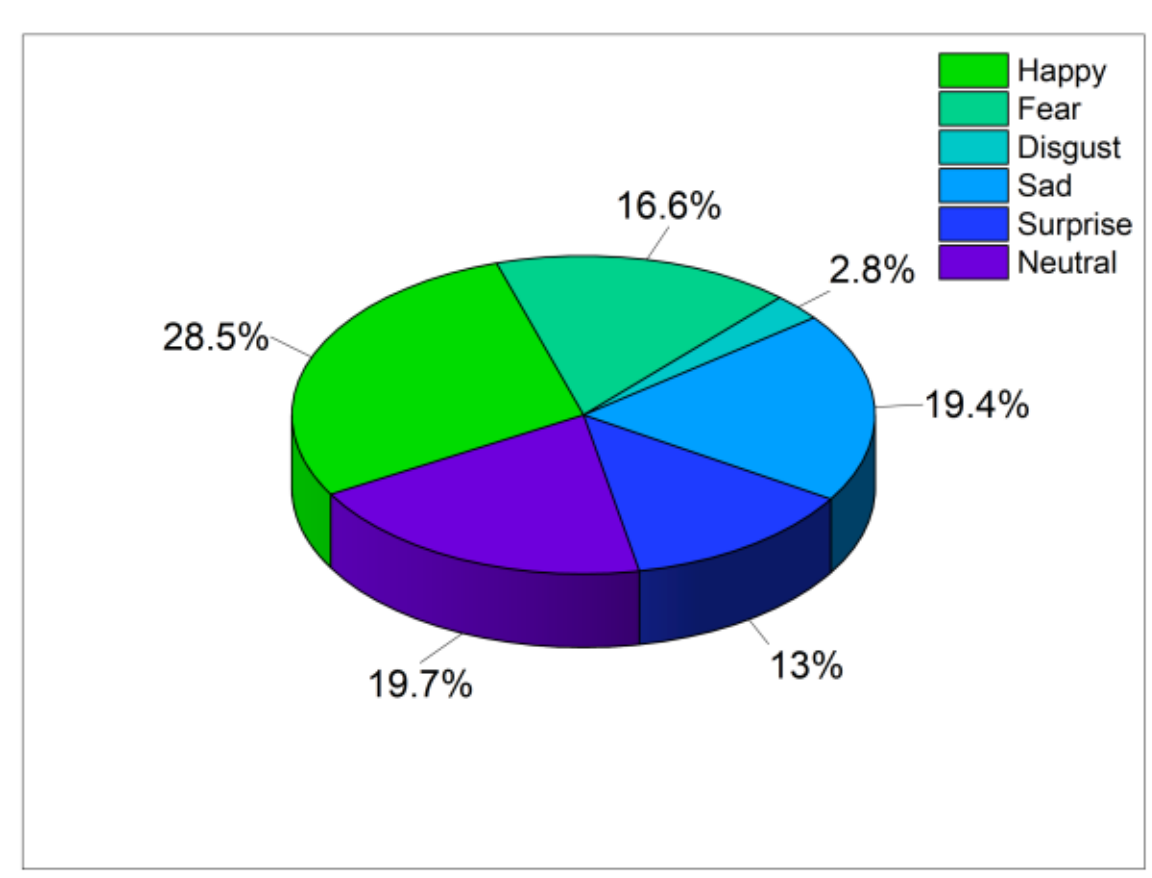

**FIGURE 1. Representation of Image Data**

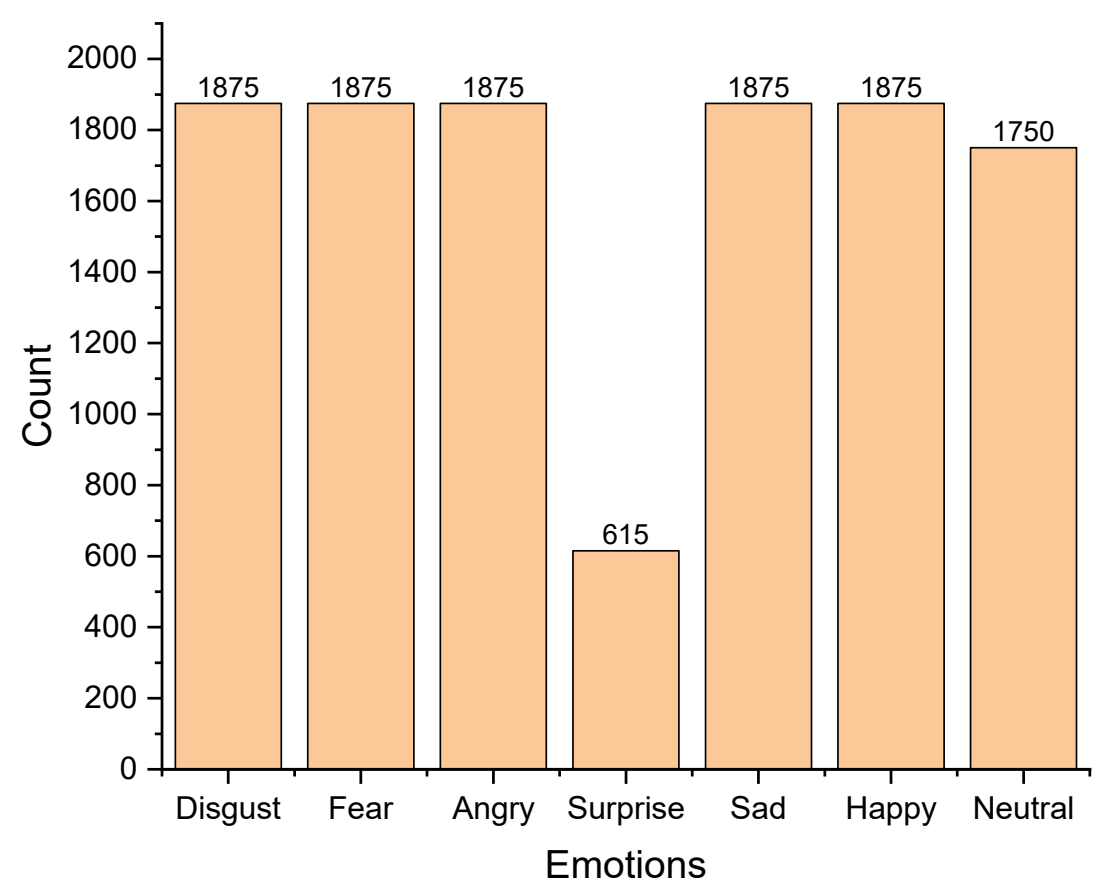

**FIGURE 2. Representation of Speech Data.**

### C. DATA CONVERSION AND PRE-PROCESSING

To use the image dataset to train the model we formatted all images since the Convolutional Neural Networks are known to work best on grayscale images. The FER dataset already had images as 48 x 48 grayscale pixels. The additional local datasets as originally shown in Fig. 3 were also further converted from their original 365 colours (RGB) pixels to grayscale images using the luminosity methods. With this approach, the intensity of red is decreased whiles the intensity of green is increased. The blue colour is moderately increased in between red and green. The output of the conversion is shown in Fig. 4.

*Grayscale = (0.3 x Red) + (0.59 x Green) + (0.11 x Blue)*

This same process could also be achieved using the average method although this might not work in all cases and sometimes converts images to black. We used this method because, in as much as balancing the dataset was necessary, the goal is to improve and reduce bias.

Using the Dlib python image processing library, we perform abstractions on the image to carefully map out features of the face. The facial landmark predictor is used to localize the structures of the face consisting of 68 facial landmarks including the eyes, nose, mouth, lips, eyebrows, and jawlines (Fig. 5). Once the features have been carefully outlined, we used the OpenCV library to select the most prominent features necessary to train the model. This included 5 features each for the left and right brows, 8 feature points for the eyes and 17 landmark points for the jawlines. The CNN local receptive fields, shared weights, and pooling layers were used to carefully extract the features.

In most macro expressions, regions of expression usually span a larger area of the face therefore, placing a central motion region allowed for a great amount of time to observe expressions as seen in Fig 5. It also clearly observed that the philtrum, which is the space between the nose, represented by 9 points, and upper lip seem to reduce across the spectrum of emotions. The entire mouth is represented by 20 points having a special influence on the type of emotion shown. It can be seen that the right image has fewer points since both the upper and lower lips are close together.

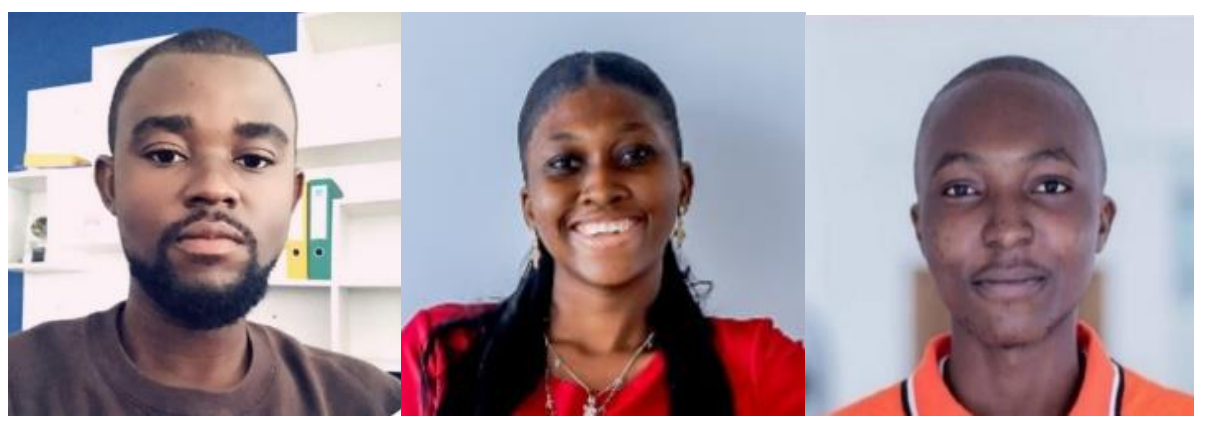

FIGURE 3. **Raw samples of the local dataset**

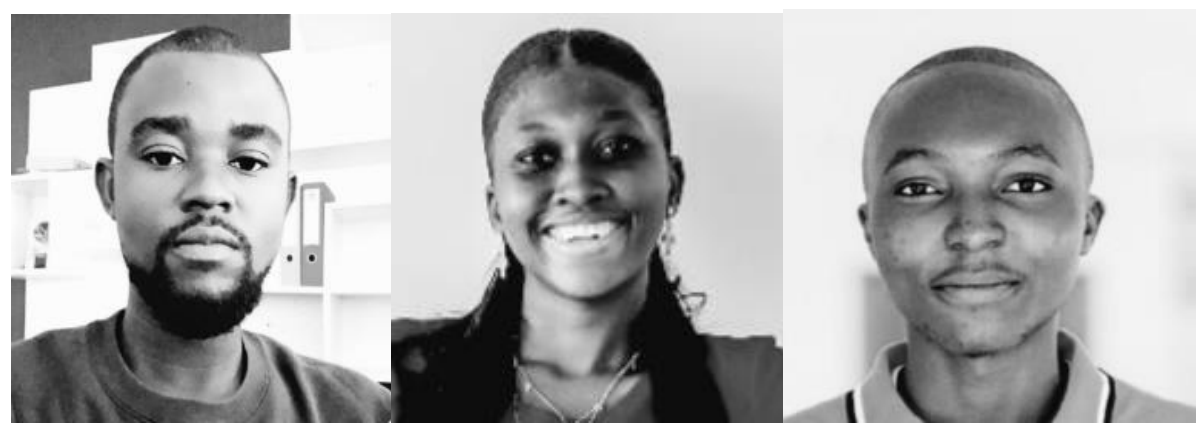

FIGURE 4. **Grayscale conversion of the dataset**

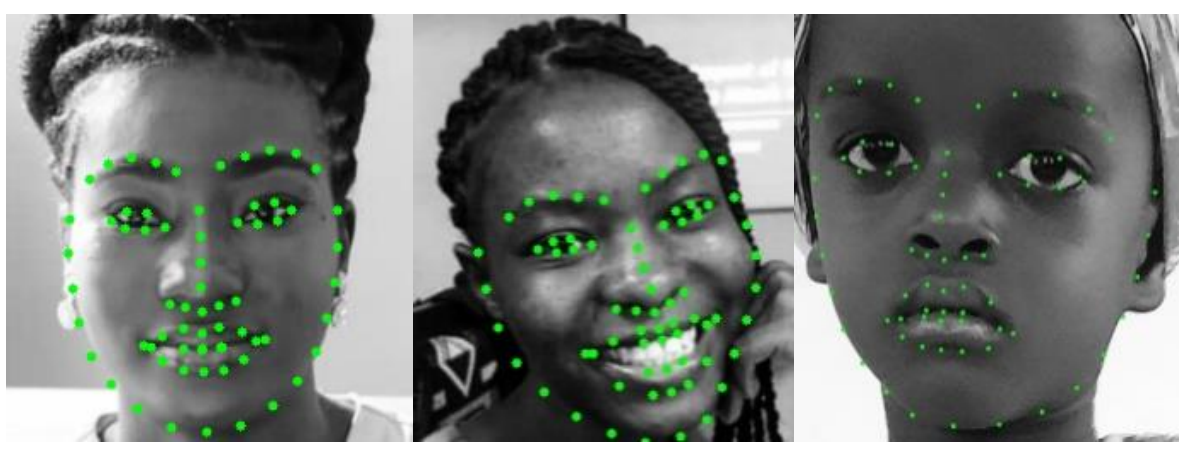

FIGURE 5. **Feature extraction**

Now, we performed the extraction of audio files which consists of two components; the sample rate and sample data. Each of the audio data samples is taken through the following steps:

- Conversion of audio to .wav file format
- Loading the audio and analyzing the frequency-time transformation.
- Feeding the audio into a Speech recognition system for transcription.
- Apply Natural Language Processing to the transcribed output for classification.

Transformation of the audio data into three-dimensional representation(amplitude, frequency and time) helps for easy transformation and computation. The amplitude-time graph is represented in Fig. 6. Also, the Fourier transform of the signal is computed using the values from both the amplitude-time graph and the frequency-time graphs as shown in Fig. 7.

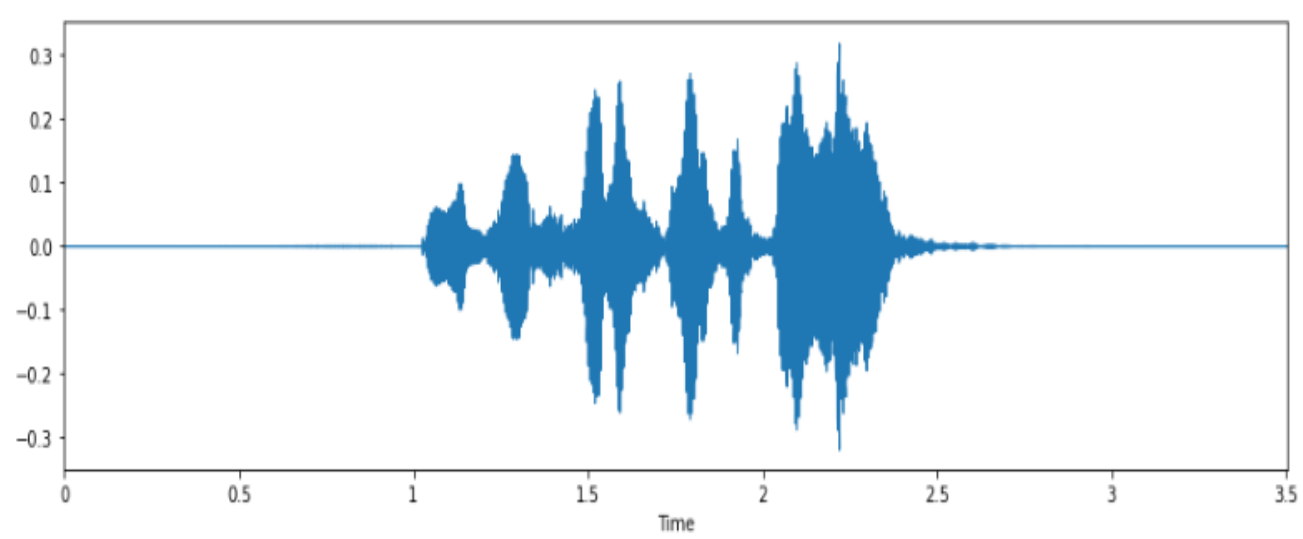

FIGURE 6. **Sample Amplitude Time Representation of audio data**

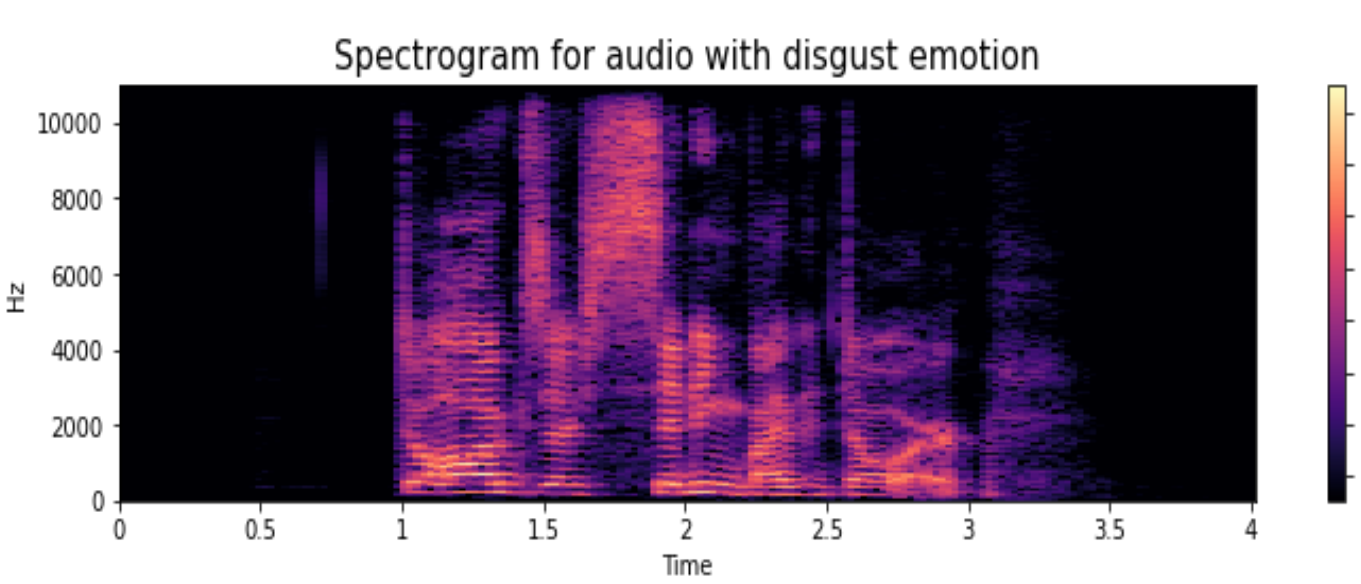

FIGURE 7. **Frequency-time graph of emotions (sample for disgust)**

## *D. PREDICTION MODEL*

This research implements the popular Convolutional Neural Network (CNN) to classify the data into their various emotions. We used CNN because it can work better on images in two dimensions without having to convert them to one dimension as in the case of using Neural Networks. Two different CNN models were used to train the image and audio datasets respectively.

Three (3) blocks of CNNS were used to train the dataset. we chose 3 blocks as ideal for our model since experimenting with the single block provided an accuracy of 72% whiles overfitting the test data. Using 3 blocks of CNN improved the accuracy significantly while addressing the rapid overfitting of the test dataset. Other regularization approaches were applied as explained in section IV. Each block contained a 2D convolutional network whose outputs were passed as inputs to a batch normalization layer. It is passed to an activation layer which in this research represents the ReLu activation function. This is then fed into a max-pooling 2D layer of pooling size 2. The final layer in the convolutional block is a dropout layer. The unit size of the initial block is 64 and this value increases by a multiple of two in the other blocks.

The VGG16 is a CNN architecture for image recognition that aims at improving the accuracy of the model. Similar to the CNN architecture with convolution and max-pooling layers, it also has two fully connected (FC) layers with a final SoftMax layer. After training image datasets to cater for predictions from image frames, we trained another model to detect emotions from the speech. The research considered two possible scenarios during the testing phase of the speech model. This includes speeches from video (ie. video to text transcriptions) and audio feed (ie. Speech from live webcams/microphones).

### *E. LIBRARIES USED IN TRANING THE AUDIO DATA*

- Zero-Crossing Rate: ZCR is the rate of change in signal from the highest value (positive) to zero mark to the least value (negative) and its reverse order in a specified time frame.
- Chroma Vector: It identifies the spectral energy using 12 binary values denoting the pitch sounds.
- Chroma Deviation: This is the variation of the Chroma Vector values from the mean which is represented by its standard deviation.
- MFCC: Mel Frequency Cepstral Coefficient is the representation of the power spectrum of the audio. This is obtained by taking the Fourier Transform of the signal and locating them unto the Mel-scale.
- Root Mean Square value: RMS is the quadratic mean which represents the square root of the mean square of the signal.
- MelSpectogram to train our model: A representation of the signal to time in a spectrum.

## IV. RESULTS AND DISCUSSIONS

### *A. CNN MODEL RESULTS*

In the beginning, a higher loss and lower accuracy are recorded in the training model as Fig. 8 shows a sturdy linear rise in the accuracy after each epoch. Meanwhile, the loss also keeps reducing which explains that the model is getting better and learning from past epochs. In Fig. 9, it can be seen that getting to the end of the 15th epoch, the accuracy rises to 65.23 per cent however a higher loss is also recorded. This means that our initial mode was not learning well and is very likely to give more predictions with errors hence the need to adjust the model parameters further.

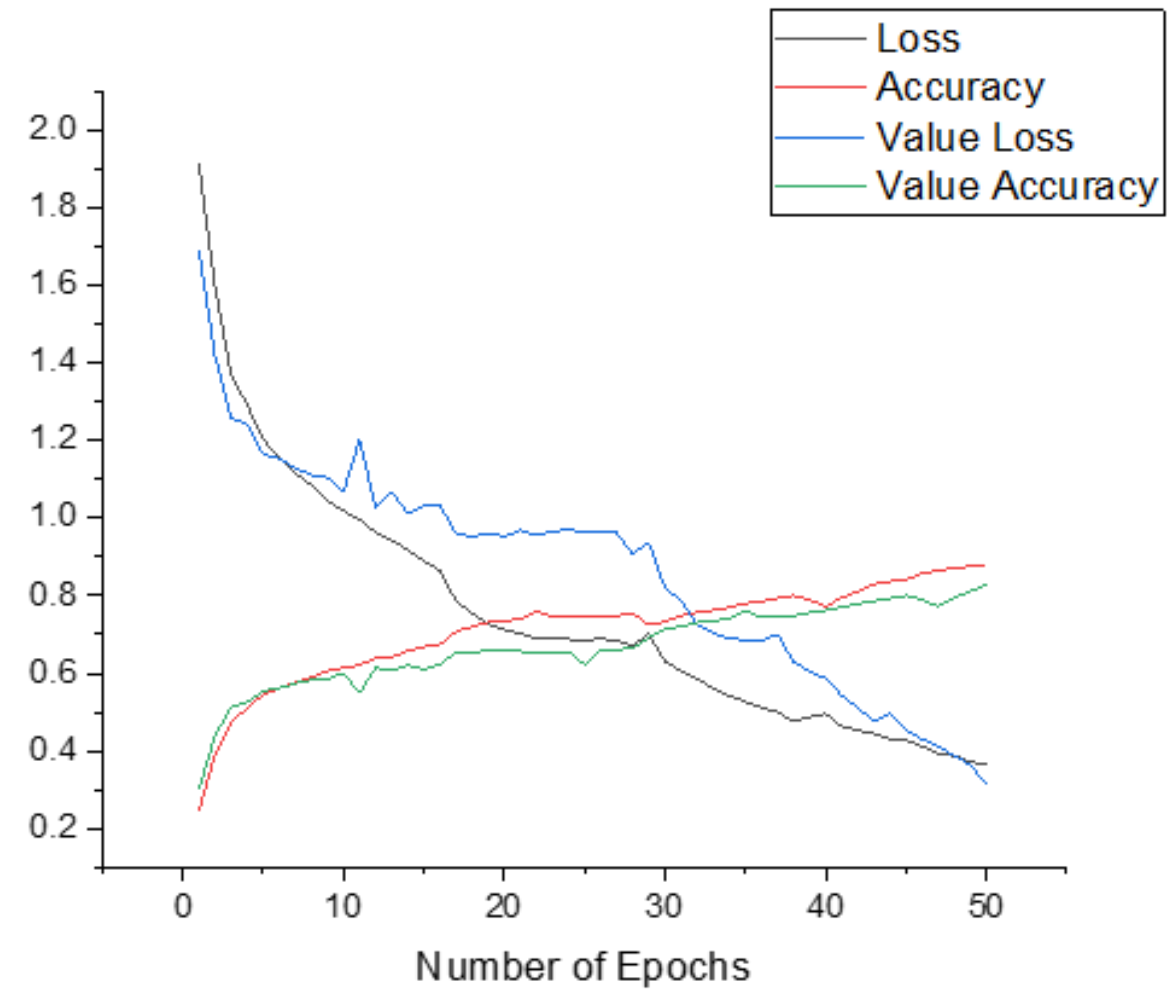

**FIGURE 8.** **Metrics after 30 epochs**

### *B. LEARNING RATE SCHEDULER, DATA AUGMENTATION, RANDOM OVERSAMPLING, AND RE-WEIGHTED*

We observed that using only the time-based, stop or exponential decay learning rates had resulted in higher accuracies but also overfitted the model. Another issue was that since the learning rate's hyperparameter values had to be pre-defined and may need to be updated with sparse data. And also, we introduced the Adam optimizer to address the learning rate and also reach global minima. A learning rate of 0.001 was used while the other hyperparameter values were left at their default. Maintaining the learning rate, while experimenting with different epoch values, it can be seen from Fig. 9 that after 30 epochs there is a steady increase in the accuracy which peaks between the 46th and 50th epoch. The loss value also drops significantly from the first 10 epochs and maintains a constant linear drop through the epochs left. This was an effect of lowering the learning rate while the decay constant was increased which reduced the momentum of the optimizer hence reducing the oscillation in the accuracy and loss values.

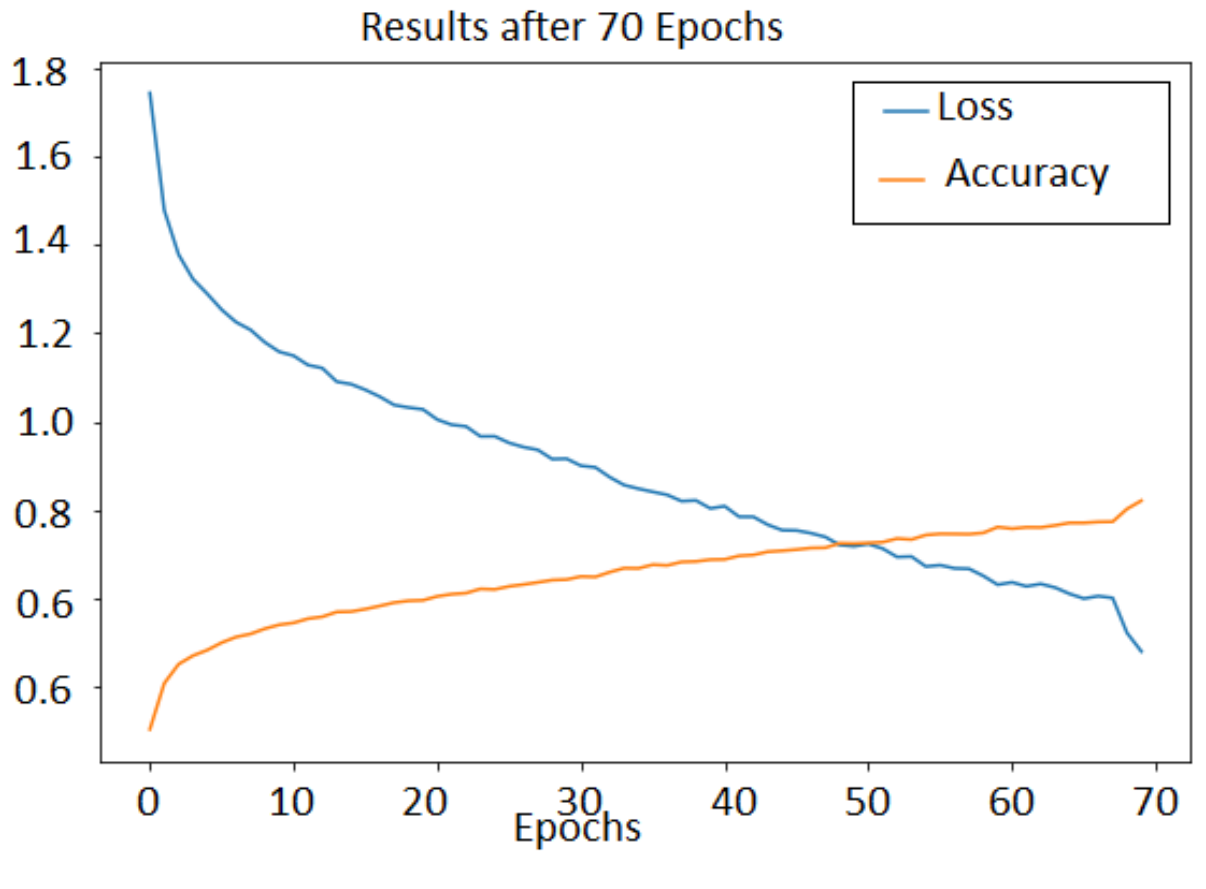

*FIGURE 9.* ***Metrics after 70 epochs***

### *C. GAP*

Difficulties in spotting micro expressions in non-static frames(videos). This was as a result of the following reasons:

- Color/Complexion of faces in contrast with the background.
- Use of facial accessories including but not limited to makeups, eyeglasses.

Increasing the number of features to train the model, as well as the number of epochs, provided a much better result as the accuracy increased to 82.59 per cent. This also had an impact on the loss value decreasing to 47.21% per cent as shown in Fig. 9. Finally, the confusion matrix (Table 1) demonstrated a clear count of the emotions detected to the actual and predicted values.

TABLE I
METRICS AFTER ML MODEL ON TEST DATA

| Emotion Class | N(Classified) | Accuracy (%) | Precision | Recall | F1 Score |
|---|---|---|---|---|---|
| Anger | 1770 | 83.18 | 0.72 | 0.83 | 0.77 |
| Disgust | 223 | 89.64 | 0.89 | 0.85 | 0.87 |
| Fear | 1203 | 85.83 | 0.87 | 0.73 | 0.83 |
| Happy | 2719 | 98.00 | 0.99 | 0.98 | 0.98 |
| Sad | 1685 | 88.38 | 0.89 | 0.81 | 0.85 |
| Surprise | 1538 | 87.34 | 0.81 | 0.90 | 0.89 |
| Neutral | 1928 | 86.01 | 0.77 | 0.79 | 0.78 |

### *D. MITIGATING FALSE POSITIVE AND TRUE NEGATIVES*

The Convolutional Neural Network (CNN) has been implemented on both facial expression recognition (FER) and speech emotion recognition (SER) datasets. The approach mixed with 20% well-labelled localized datasets to be able to cater for tests on video frames and micro-expression is proposed. In addition to the CNN implementation, a new mathematical approach termed the Audio Frame Mean Expression (AFME) is used to validate and improve the result from the neural network.

Due to the nature of the dataset described in section 3 above, some emotions such as Fear, Disgust and, Surprise were misclassified especially by using the algorithm. For example, from Table 1, some elements of sadness were predicted as neutral, fear, and disgust. This number of misclassified emotions constituted close to one-third of the total number of fear test data. There were also some occurrences of edge cases whose emotions lay between two contrasting emotions.

To mitigate such misclassifications, this research introduced the Audio-Frame Mean Expression. This algorithm is to validate emotions since we were considering both image frames and audio This simply means, in each video to be classified, sub videos in the range of 5-8 seconds were taken. These sub videos were taken in such a range to avoid missing out on micro-expressions. Out of the sub videos, each of the unique expressions identified using the ML algorithm was represented as discrete random variables, X. Three values of the variable X at which the probability mass function takes the maximum values are stored from both the image (variable A) and audio classifications (variable B). Now, the algorithm relies on the emotion spectrum stipulated by Plutchik [30] and assigns weights to the emotions. This spectrum enables the algorithm to predict near values of emotions which lies in between two primary emotions on the wheel. Different classifications from both videos and audios which were at the extreme ends of the wheel denoted sarcasm.

### *E. PSEUDOCODE FOR THE AUDIO-FRAME MEAN EXPRESSION*

**Algorithm 1**: The algorithm of the audio-frame mean expression
**Input**: Multiple emotions A and B with their percentage matches
**Output**: The emotion of the audio and image process completion

```
1  M – individual emotions detected from speech data with
     their percentages as dictionaries A;
2  N – individual emotions detected from image data with
     their percentages as dictionaries B;
3  detect – false;
4  for each i = M, N do
5      for each m = 1,2, …., M, do
6          Obtain the 3 highest scoring emotions
           B = {p1, p2, p3};
7      end
8      for each n = 1, 2, . . ., N do
9          Obtain the 3 highest scoring emotions
           C ={p1, p3,p3};
10     end
11 end
```

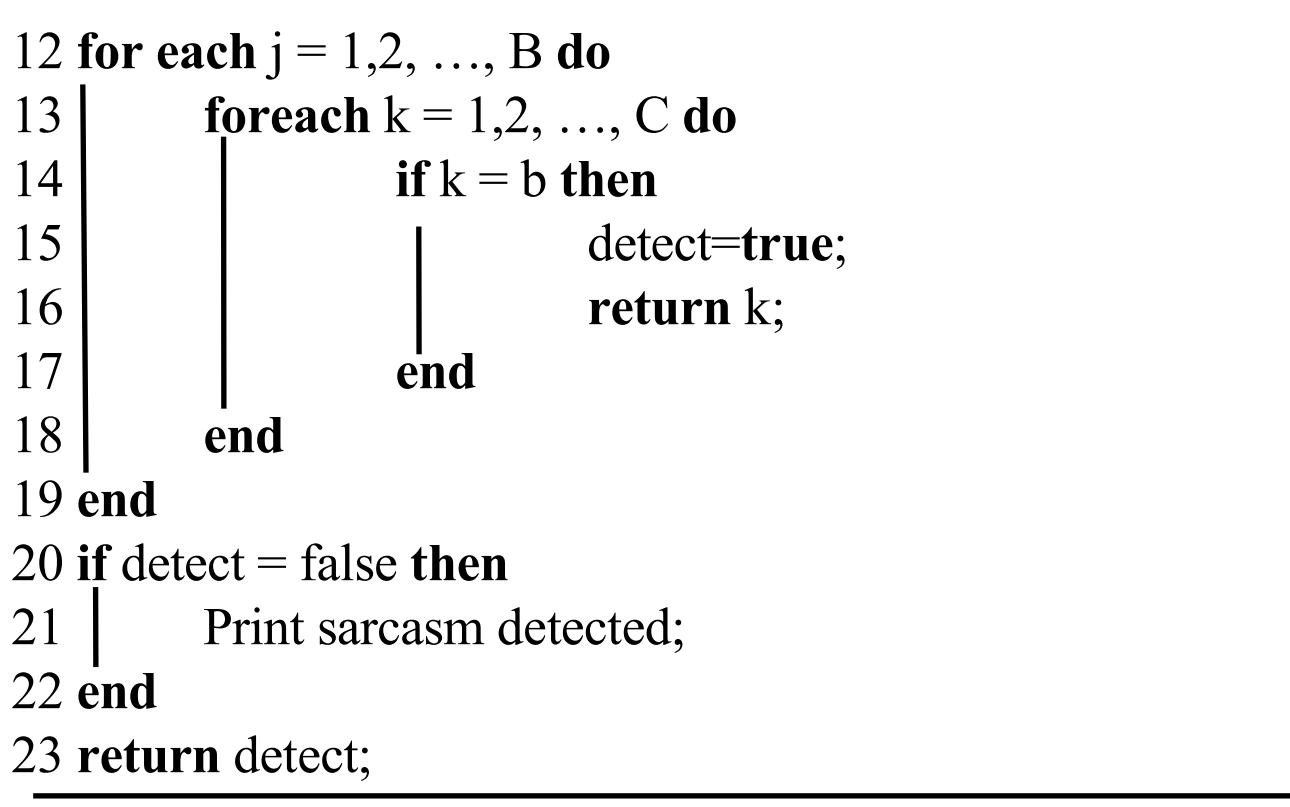

```
12 for each j = 1,2, …, B do
13 |    foreach k = 1,2, …, C do
14 |    |    if k = b then
15 |    |    |    detect=true;
16 |    |    |    return k;
17 |    |    end
18 |    end
19 end
20 if detect = false then
21 |    Print sarcasm detected;
22 end
23 return detect;
```

### F. ANALYSIS AFTER AUDIO-FRAME MEAN EXPRESSION

For example, we tested our model on a public youtube video of a Nigerian writer [31]. Fig. 10 shows the captured micro frames within the 5th-6th seconds and the words spoken in that interval. The happy emotion was expressed within this interval however the extracted audio ("of course it was angry") was classified as Anger. Using our AFME algorithm we were able to extract these emotions and compared them on Plutchick's wheel of emotion [30]. After, comparison and identifying that the separate emotions were extremely contrasting, the algorithm hence detected sarcasm.

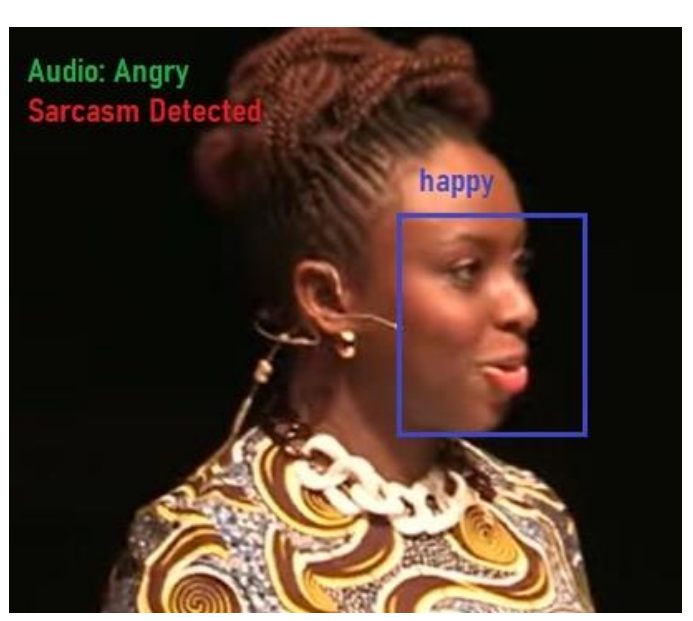

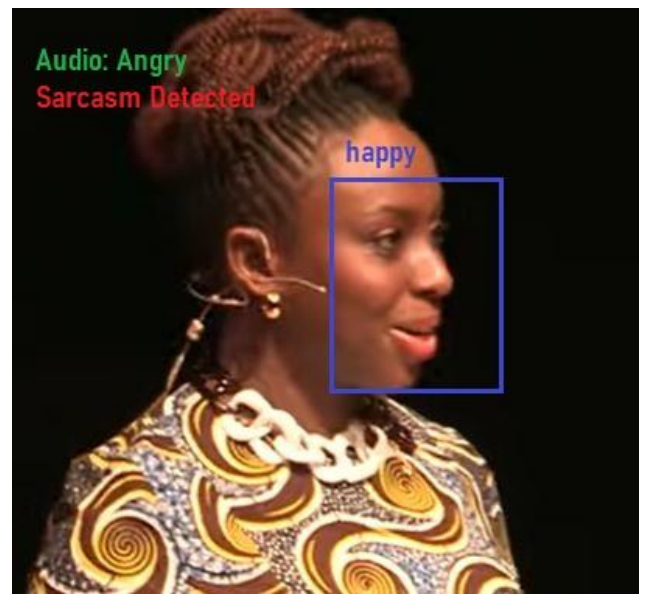

**FIGURE 10.** *Sample emotion detection on video frames using our model*

TABLE 2
PERFORMANCE METRICS AFTER AUDIO-FRAME MEAN EXPRESSION

| Emotion Class | N (truth) | N (classified) | Accuracy (%) | Precision | Recall | F1 Score |
|---|---|---|---|---|---|---|
| Anger | 1531 | 1770 | 93.18 | 0.72 | 0.83 | 0.77 |
| Disgust | 283 | 243 | 94.64 | 0.96 | 0.85 | 0.92 |
| Fear | 1581 | 1203 | 85.83 | 0.97 | 0.73 | 0.83 |
| Happy | 2719 | 2868 | 100 | 1.0 | 0.94 | 0.96 |
| Sad | 1846 | 1785 | 97.34 | 0.97 | 0.91 | 0.94 |
| Surprise | 1244 | 1238 | 98.38 | 0.81 | 0.93 | 0.86 |
| Neutral | 1882 | 1978 | 95.80 | 0.87 | 0.89 | 0.88 |

TABLE 3
COMPARISON OF FINAL RESULTS WITH OTHER MODELS

| | Model | Dataset | Accuracy |
|---|---|---|---|
| AFME (this paper) | CNN + AFME | FER 2013, Ravedess, Savee, Tess, Local Data | 95.7 |
| Syed [32] | Mini Xception | FER 2013 | 95.6 |
| Giannopoulos [20] | CNN | FER 2013 | 83 |
| Dachapally [19] | Autoencoder | JAFFE | 86.38 |
| Jaiswal [21] | CNN | FER 2013 | 70.14 |
| Khaireddin [22] | CNN | FER 2013 and JAFFE | 73.28 and 98 |

This approach does not only seem to improve the outcome of the classification but is also an effective and optimized way considering the number of variables (emotions) present. Table 2 compares the output of the recall, precision, and F1 scores of each of the experiments performed based on the values recorded for the True positives, true negatives, false positives, and false negatives. It can be observed that the accuracies improved as compared to similar research (Table 3). However, based on the precision and recall values, we noticed a small number of misclassifications which means that the introduction of the AFME provided better predictions on the test data and also detected sarcasm.

### G. REASON FOR CONSIDERING BOTH SPEECH (AUDIO) AND IMAGE FRAMES ASSOCIATED WITH SPEECH

#### 1) SARCASM

Sarcasm may have different text matching a particular facial expression also known as visual paralinguistic [33], therefore cannot be overlooked [11]. Most often, context could be used to detect sarcasm in cases where there is a slight or vast difference relating to the subject matter. The tone and gestures expressed while speaking could create cues to detect whether a particular emotion is characterized with humour or irony.

#### 2) RATE OF SPEECH

Automatic tagging of prosodic phrases which are segments of speech that occur with a single pitch is employed to extract natural language data from real-life conversations using the support vector machine model [12]. This approach aided in the segmentation of words and phrases including non-verbal content. Achieving a 52% accuracy using this Natural Language Processing (NLP) model, the result showed that even with the seven major emotions, non-verbal sounds played a crucial part in achieving high accuracies for emotion detections.

#### 3) NON-VERBAL COMMUNICATION AND CUES

Non-verbal communication is a huge part of our communication kit. This is where body language or facial expression is used while speech is minimal or non-existent.

Recently, due to the introduction of text messaging-based software such as WhatsApp, telegram, and Facebook, people have found ways to express sentiments and tone within the text using emoji. This creates a big difference when an individual receives a text saying “okay” and an “okay :)”. In as much as there might be cases where wrongful usage of emojis may lead to a misleading interpretation, a person’s psychometric characteristics can be measured using self-identification with emojis.

### H. RECOMMENDATION ON STRUCTURE OF DESIGNING CONVERSATIONAL AGENTS

- Conversational agents need to be designed to strongly meet the following criteria:
- Perceive emotions and surroundings.
- Respond to emotion and communicate with high-level dialogue.
- Recognize emotion recognition models of other agents.
- Create social relationships and maintain them to remain competent - learn social qualities leading to trust.
- Depending on natural cues.

## V. CONCLUSION

Limitations such as access to African data might take some time to be solved. Until there can be access to a readily available well-structured dataset on the emotions of Africans, there needs to be algorithmic and dataset restructures. By balancing the datasets, and adjusting our new Audio-Frame Mean Expression, we designed a better model to predict emotions on black demographics. We relied on the use of CNN to provide good accuracies for the classification of the model on both speech and text data. From our model results, we identified that the standalone results are not enough to accurately predict the emotions on black faces. The performance analysis of the results before and after the mitigating techniques were applied showed an improvement in the F1-score, Precision, and Recall of the model. Our goal here is to increase the number of true positives and false negatives while strategically reducing the number of false positives and true negatives. We also identified that the implementation of other traditional mathematical techniques to the modern machine learning models would help mitigate false positives and true negatives.

## VI. FUTURE WORK

For future research, the dataset used could be first subjected to IBM 360’s model [23] to identify other areas of bias within the dataset. We propose an implementation and comparison of different mitigation techniques proposed by the IBM 360 model to find an optimal Emotion Recognition model. Although this method might be tedious and time-consuming, it serves to also validate the causes of bias mentioned in this paper.

Secondly, more attention could be paid to the different features that classify faces and voices amongst different groups of people. This could be introduced as a new parameter for model detection.

## ACKNOWLEDGMENT

We would like to acknowledge Auxane Boch, a researcher specializing in the Ethics of AI at the Technical University of Munich for her contributions.